\documentclass[conference]{IEEEtran}
\usepackage{times}
\usepackage{helvet}
\usepackage{courier}
\usepackage{url}
\usepackage{graphicx}

\usepackage{filecontents,lipsum}
\usepackage[noadjust]{cite}
\author{Joel Michelson, Joshua H. Palmer, Aneesha Dasari, and Maithilee Kunda\\
Electrical Engineering and Computer Science, Vanderbilt University, Nashville TN, USA
}

\graphicspath{ {usedImages/}}
\usepackage{enumitem}
\usepackage{amsmath}
\usepackage{algorithm}
\usepackage[noend]{algpseudocode}

\title{Learning Spatially Structured Image Transformations Using Planar Neural Networks}

\begin{document}

\maketitle

\begin{abstract}

Learning image transformations is essential to the idea of mental simulation as a method of cognitive inference.  We take a connectionist modeling approach, using planar 
neural networks to learn fundamental imagery transformations, like translation, rotation, and scaling, 
from perceptual experiences in the form of image sequences. 
We investigate how variations in network topology, training data, and image shape, among other factors, 
affect the efficiency and effectiveness of learning visual imagery transformations, including effectiveness of transfer to operating on new types of data.
  
\end{abstract}

\section{Introduction}

Visuospatial reasoning is ubiquitous in everyday human intelligence.  In addition to its reliance on \textit{semantic} knowledge about objects, categories, and scenes, visuospatial reasoning also requires \textit{non-semantic} knowledge about object shapes, spatial relationships, etc., including, for example \cite{newcombe2015thinking} (p. 182):
\begin{quote}
``Transforming the spatial codings of objects, including expansions or reductions in size, rotation, [etc.]...accumulating  sequences  of  such  changes and visualizing change over time....''
\end{quote}

\noindent We do not know exactly how the human brain represents such non-semantic visuospatial knowledge about transformations, but we do know that this knowledge is learned through real-world perceptual experiences, especially in infancy and early childhood \cite{smith2005development}; and that it is often deployed through top-down neural activations in brain regions associated with visual perception, i.e., using visual mental imagery \cite{pearson2015heterogeneity}.

Only a few studies have examined how AI systems can represent and learn transformation-based reasoning operations like image rotation from perceptual experience. One early study represented each operation as a distributed set of weights in a single-layer, 2D connectionist network, and used the perceptron learning rule to learn each operation in a supervised fashion from image sequences depicting that operation \cite{mel1986connectionist}.  

Other recent work uses similar distributed representations of operations but adds a hidden layer to the network that enables learning multiple operations in an unsupervised fashion \cite{memisevic_unsupervised_2007,memisevic_learning_2010}.  Another study uses a robot architecture combining visual and motor inputs for learning the rotation operation \cite{seepanomwan2013modelling}.

(Note: learning an operation like rotation is \textit{not} the same as learning transformation-invariance (e.g., for object recognition), though the two may be used in conjunction for certain tasks.  These are also distinct processes in human cognition \cite{farah1988mental,vanrie2002mental}.  Thus, AI research on representing and learning transformation invariance \cite{foldiak1991learning,anselmi2016unsupervised} is \textit{not} directly relevant to the work presented here, though some recent research combines both ideas to address problems in computer vision \cite{jaderberg2015spatial}.)

In this paper, inspired by the simple and effective approach proposed by Mel \cite{mel1986connectionist}, we examine how planar neural networks---2D arrangements of perceptrons that can pass visual information to neighboring units---can represent various image transformations such as translation, rotation, and scaling.  We also investigate how differences in learning algorithm parameters, training data, and spatial network layout and topology affect learning performance.

More broadly, learning image transformations as functions that can operate over novel visual material is essential to the idea of mental simulation as a method of cognitive inference.  Ultimately, such representations could be added to AI systems to support robust visuospatial reasoning, and will also help improve our understanding of how humans represent and learn mental image transformations.  

\begin{figure}[t]
    \centering
    \includegraphics[width=0.8\linewidth]{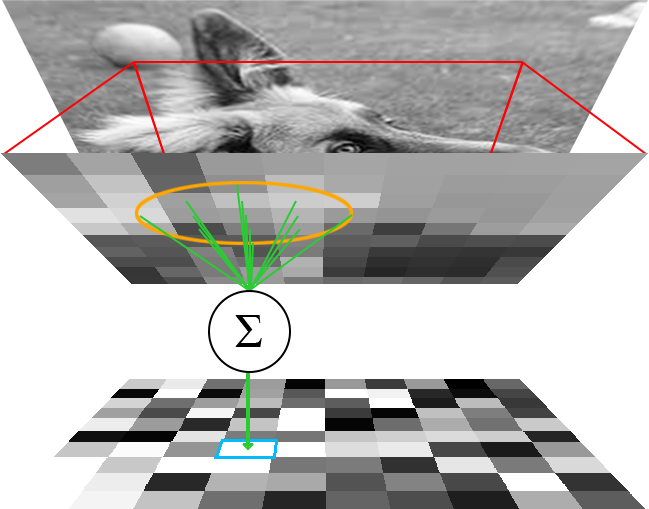}
    \caption{A real world image is sampled from, perceptrons take input from euclidean neighborhoods (in this case $radius=2$) of that sampling, and then each perceptron outputs to one pixel in the resulting image}
    \label{fig:grid}
\end{figure}

\section{Methods}

We use a supervised learning approach.  Given a pair of training images depicting some transformation (e.g., rotation having a particular direction and velocity), each perceptron in our plane-of-perceptrons neural network learns how the visual information at its own location in the visual field is transformed.  \textbf{Thus, each transformation is represented as a set of local pixel intensity flows distributed across spatial locations in the network.}  This is similar to notions of optical flow, except that information about a given transformation is learned using purely local visual information.  No explicit feature matching is done across input images. Each perceptron in the network learns independently from the rest.



Our system is arranged as a plane of perceptrons, each with a \textit{position} label given by Cartesian (or Polar) coordinates. Given adjacency function $U$, node $p_{0}$ has an incoming edge from node $p{\iff}p{\in}U(p_{0})$. By default, $U(p_{0})=U(pos(p_{0}),2)$, the open ball of radius 2 centered on $p_{0}$. 


Given input image $X$, node $j$, and its bias $b_j$, the network output of node $j$ is as follows:
\begin{center}
$out_{pos(j)} = \sigma(b_j+\sum\limits_{i{\in}U(j)} w_{i\rightarrow j}*X_{pos(i)})$
\end{center}

In an image, pixel intensities are valued between 0 and 1, so we clamp each perceptron's output likewise:

\begin{center}
$\sigma(x) = max(0, min(1,x))$
\end{center}

A given pixel in the output image corresponds to the output of the node at that pixel's \textit{position}, so that pixel's value is determined by a linear combination of the values of the neighboring input pixels.

Weights between nodes are learned using the perceptron update rule: given nodes $i$ and $j$, target (transformed) image $T$, network prediction image $P$, and learning rate $\delta$,

\begin{center}
${\Delta}w_{i\rightarrow j}=\delta(T_{pos(j)}-P_{pos(j)})X_{pos(i)}$
\end{center}

Weights are updated after a batch by the sum of all such deltas from images in the batch.

\textbf{Data Generation.}
Transformations can create blurring and information loss; thus we transform both the inputs and target outputs exactly once as follows. 

A high resolution image $I$ is generated or loaded. Let $N$ be the network. Let $T$ be the transformation we want $N$ to learn and $R$ be a random initial transformation of the same type. Then we can calculate $I_j:=T^j \circ R(I)$ for $j{\in}\{0,1,2,5\}$. Each $I_j$ is downscaled to the dimensions of $N$ via bilinear sampling. $I_0$ is the input. If $I$ is in the training set, $I_1$ is the target for training. Otherwise, $I_1$, $I_2$, and $I_5$ are targets for testing where error is evaluated as $|N^j(I_0)-I_j|$, the mean of the absolute errors of all the pixels in an image. We also call these images 1X-, 2X-, and 5X-chained outputs.



\section{Illustrated Examples}

Before diving into experiments, we provide a detailed demonstration of the  network's ability to learn three transformations: translation, rotation, and scaling. 
Our network's inputs and target outputs are pairs of 16x16-pixel samplings from 160x160 centers of 320x320 images. To reduce the effect of the law of large numbers, we generate random noise at a lower resolution than our `real world' images and upscale by a factor of 10.  Transformations are applied, and the resulting images are downscaled bilinearly to size 16x16.


For each of these demonstrations, we use the same hyperparameters: a learning rate of 0.01, a neighborhood with a constant radius of 2, an image size of 16x16, a batch size of 50, 250 training image pairs, and 50 testing image pairs. Weights are initialized as random values between -1 and 1. We train for 40 epochs.

We display network structures using blue and red graphs. Blue lines indicate positive weights and red negative. To reduce clutter, opacities scale with the squares of weights, and weights with magnitudes less than 0.2 are omitted.

\begin{figure}[!t]
    \centering
    \includegraphics[width=\linewidth]{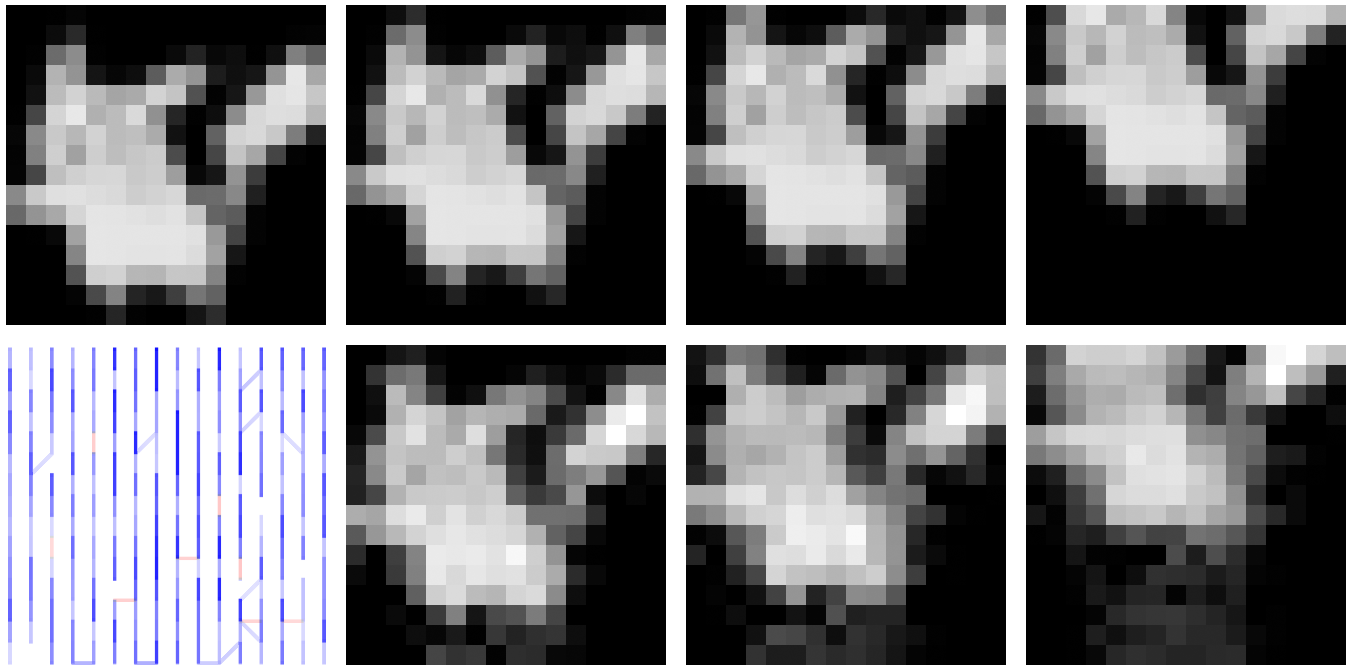}
    \par
    \vspace{9pt}
    \includegraphics[width=\linewidth]{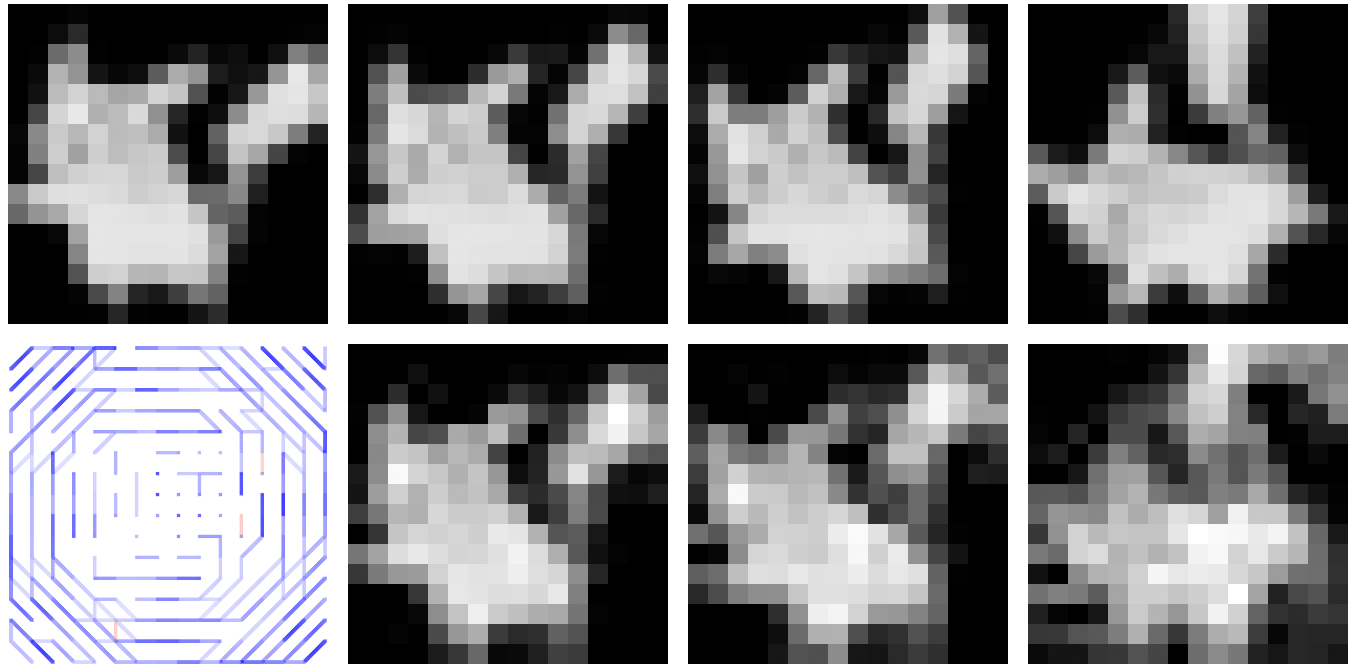}
    \par
    \vspace{9pt}
    \includegraphics[width=\linewidth]{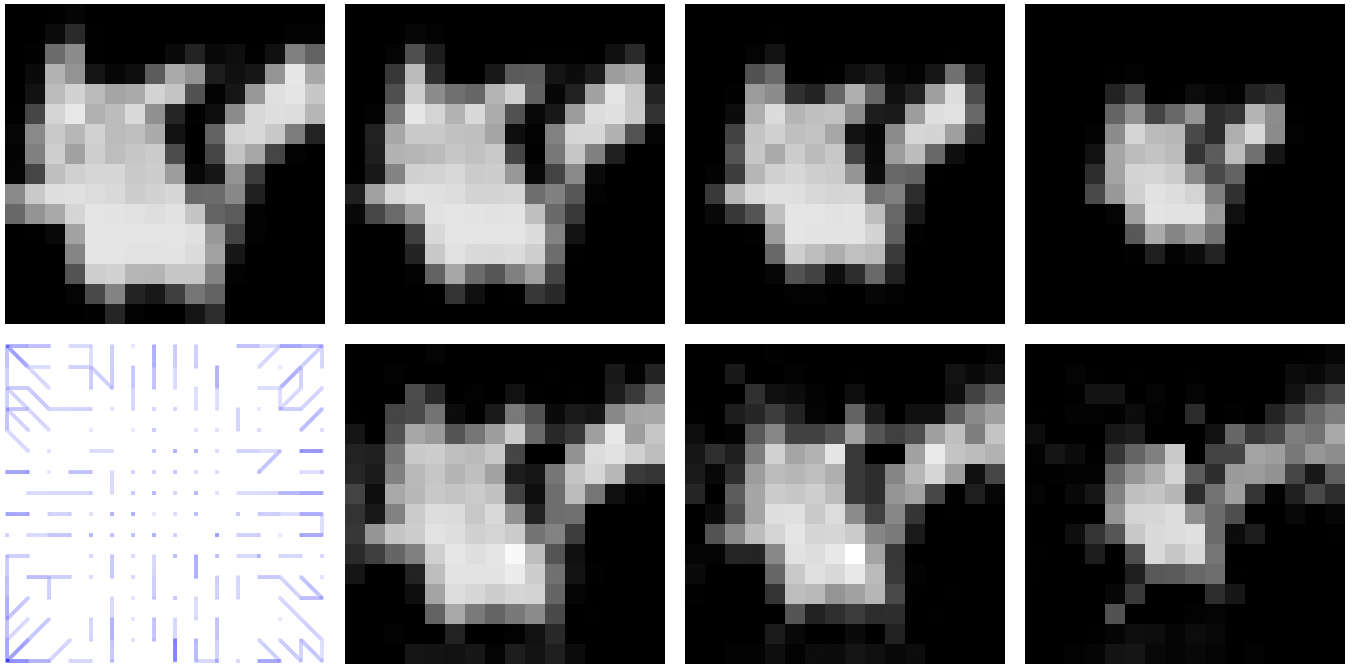}
    \caption{Illustrative results for translation (top), rotation (middle), and scaling (bottom).  Top row of each section shows training images for 1X-, 2X-, and 5X-chained transformations.  Bottom row of each section shows network structure after training (left), followed by network outputs for three chain lengths.
    }
    \label{fig:baseTranslate}
\end{figure}

%

    


\textbf{Translation.}  
We shift the original images by (0,1), or one pixel upwards. 
Since our network is rectangular, neurons away from the edge of the image are be able to perform translation perfectly, provided a sufficient neighborhood size and integer-length translations. Unlike many machine learning problems, we know exactly what the weights should be: 1 if connecting a pixel to its neighbor above it, and 0 otherwise. Translation is perfectly represented, so chaining translation is as well.

\textbf{Rotation.}
We rotate the original images ten degrees counter-clockwise. Unlike translation, it is impossible for the rectangular network to perfectly replicate rotation since edges with a neighborhood size of 2 may only cover 8 angles (0, 45, 90, etc.) of three lengths (0, 1, 1.414). There are only nine regions of high accuracy in the in any trained network's output, each corresponding with an almost correct angle/length combination. As such, training and testing errors will never converge to 0.



\textbf{Scaling.}
We scale the original images by a factor of 0.9 about the center of the image. Like rotation, scaling is not perfectly covered by a rectangular array of neurons, so the average error for scaling cannot approach 0. 


\begin{figure}[t]
    \centering
    \includegraphics[trim={0 0 0 1.5cm},clip,width=\linewidth]{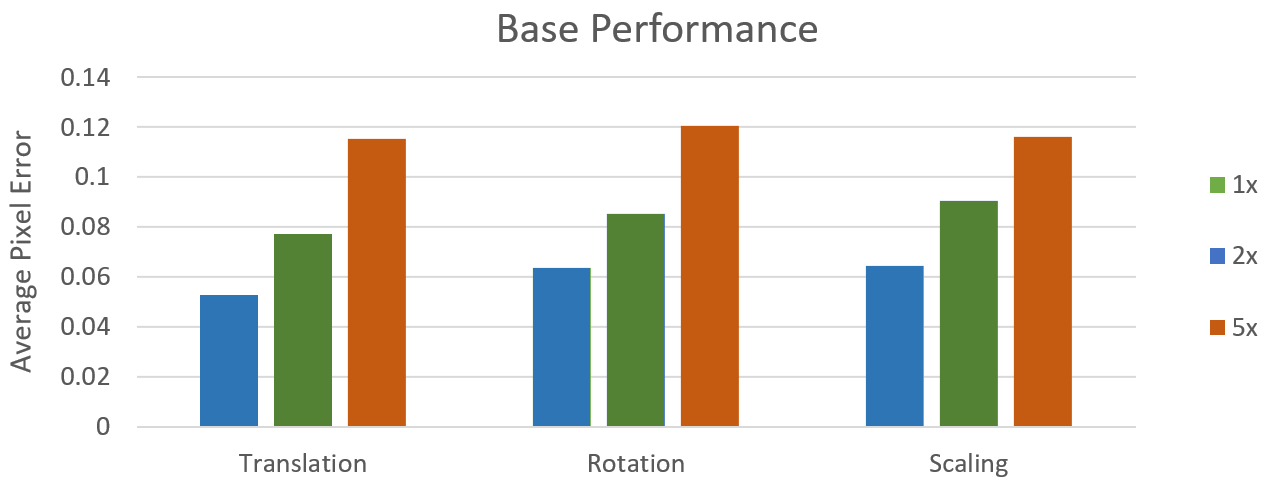}
    \caption{The average pixel errors for all three base transformations on 1X-, 2X-, and 5X-chained test sets}
    \label{fig:base}
\end{figure}

Figure \ref{fig:base} summarizes the performance of these three transformations. Translation outperforms at first; however due to the chaining of aforementioned edge errors, 5X translation loses its edge. Rotation only loses information in the corners, whereas scaling cannot predict values around the entirety of its borders. Thus, rotation performs slightly better.

\section{Initial Hyperparameter Search}

We demonstrate the effect of different batch sizes on rotating rectangular images. Four sizes are tested: 12, 25, 50, and 100. Each test is run for 40 epochs.

\begin{table}[h]
\caption{Average test set error with differing batch sizes}
\centering
\begin{tabular}{lrrrr}
Batch Size & 10     & 25     & 50  & 100       \\ 
\hline
200\textsuperscript{th} Batch Error & 0.152     & 0.096     & 0.063  & N/A     \\ 
\hline
40\textsuperscript{th} Epoch Error & 0.107 & 0.072 & 0.063 & 0.306 \\
\end{tabular}
\end{table}

Next, we test the effects of using four different learning rates during training: 0.001, 0.005, 0.01, 0.02, and 0.04. The greater two learning rates converge to high error rates rather quickly, but the smaller three perform similarly. 0.001 and 0.005 appear to follow nearly the same curve as 0.01, but they train around a tenth and a half as quickly, respectively

Since the goal of this paper is not to maximize results but to analyze the reasoning behind network errors, we do not exhaustively search for the optimal batch size or learning rate. Instead, we hope to demonstrate our network's robustness to minor changes due to the suitability of our dataset. For both variables, increasing the number incrementally speeds up training, but exceeding some value causes accuracy to plateau. A network with a batch size of 50 and a learning rate of 0.01 performs reasonably well, so we use these values in further experiments.


\section{Varying Image Generation Parameters}

\textbf{Rotation Degrees.}
We trained our network to perform four rotations: 5, 10, 15, and 20 degrees. 
Nine regions of low error are arranged in a square, each corresponding with one of the nine possible edges in the neighborhood with a radius of 2. Outside the square, no edges are long enough to aid in output prediction. Increasing the degree of rotation decreases the size of this square since the distance pixels must travel increases. Increasing the neighborhood size adds more low-error regions at the cost of requiring a lower training rate.


\textbf{Discrete versus Continuous Translations.}
The comparison of most interest is discrete vs. `continuous' translations, so we compare the training of a network learning to translate with an argument of (1,1) and one learning to translate with an argument of (0.5,0.5). After training, the discrete transformation reaches a test set error of 0.056 and the continuous transformation reaches an error of 0.071. Their errors remain similar even when chaining.

Although the discrete transformation performs better than the continuous transformation, the difference between their errors is minor. The continuous transformation has an advantage: its pixels travel half as far, meaning it suffers much less from the effect of offscreen pixels moving onto the image. This advantage is increased with chaining, causing error metrics to report that the continuous translation is more robust than it truly is.

\section{Varying Network Connectivity}

\textbf{Neighborhood Size.}
Increasing the radius of the neighborhood allows for more extreme image transformations to be captured but also increases the number of parameters. We train on rotation using six radius lengths from 1 to 3. 1-radius nodes receive input only from their positions.

\begin{table}[t]
\caption{Results of training with differing neighborhood radii.}
\centering
\begin{tabular}{lrrrrrr}
Radius & 1     & 1.41     & 2  & 2.24     & 2.83  & 3  \\ 
\hline
Nbhd & 1     & 5     & 9  & 13     & 21  & 25  \\ 
\hline
Error  & 0.14 & 0.08 & 0.06 & 0.07 & 0.30 & 0.33 \\
\multicolumn{7}{l}{$Nbhd$ refers to number of nodes in a neighborhood.}\\
\end{tabular}
\end{table}

\begin{figure}[t]
    \centering
    \includegraphics[width=8.4cm]{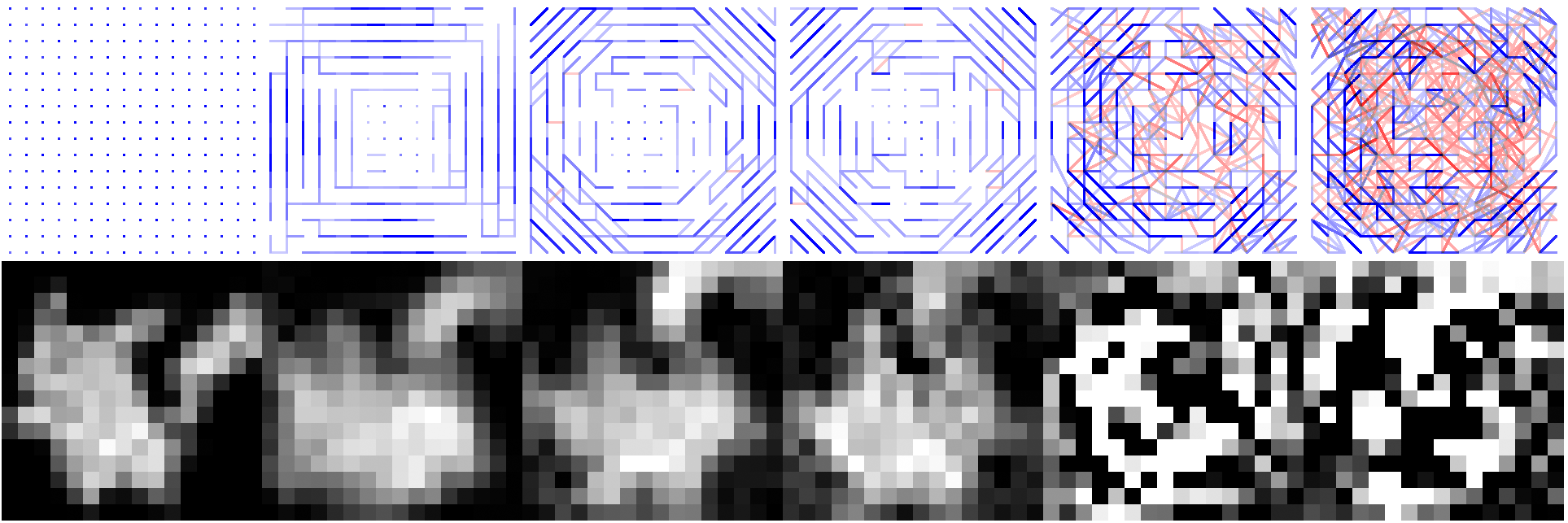}
    \caption{Top: Networks trained on 10-degree rotation with the 6 different neighborhood radii. Bottom: 5X-chained Pikachu outputs to display errors of the above networks.}
    \label{fig:nbhd}
\end{figure}

A radius of only 1 reaches a plateau of accuracy quickly as it cannot move pixels in an image. It learns to output almost the same image as the input, as seen in Figure \ref{fig:nbhd}. 
Radii of 1.41, 2, and 2.24 all train similarly to one another, but the networks with larger radii reach lower accuracy plateaus. The large number of edges in each neighborhood causes the output to wildly fluctuate between 0 and 1. Use of a sigmoid function in the perceptrons' outputs might alleviate this problem. We perform an additional run of the 3-radius network with a learning rate of 0.005. The new network does not plateau as does the same network with a larger learning rate, attaining an error of 0.138 after 200 batches.

\section{Transfer Learning}

\textbf{Various Noise Techniques.}
When images randomly generated at high resolutions are downscaled to fit our network, values are near 0.5 due to the law of large numbers. Thus we generate noise at low resolutions and then upscale for transformations. Here we demonstrate the effects of using upscaled versus high resolution noise.

\begin{figure}
    \centering
    \includegraphics[width=8.4cm]{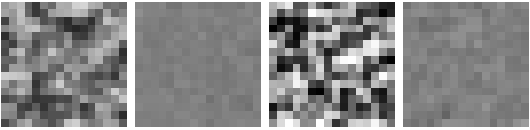}
    \caption{Sample training images: Random Noise, High-res Random Noise, Random Dot, High-res Random Dot}
    \label{fig:upscaling}
\end{figure}

\begin{itemize}
  \item \textit{Random Noise:} 32x32 images are generated, each pixel valued between 0 and 1, and then upscaled to 320x320
  \item \textit{High-res Random Noise:} 320x320 images; each pixel is a value between 0 and 1
  \item \textit{Random Dot:} 32x32 images generated and then upscaled to 320x320; each generated pixel is 0 or 1
  \item \textit{High-res Random Dot:} 320x320 images; each pixel is a value of 0 or 1
\end{itemize}

The high-res group learns to output 0.5 for each pixel, converging to test error of nearly zero after only ten batches. Both network structures in the high-res group have seemingly random connections, and the structures in the low-res group learned connections indicative of rotation.


Because the networks' errors can be low when they decidedly fail to learn rotation, solely relying on test error as a performance metric is problematic when comparing across datasets. Instead, we must measure how well learning transfers across differently generated datasets to discern whether a network performs the datasets' shared transformation.

\begin{figure}
    \centering
    \includegraphics[width=8.4cm]{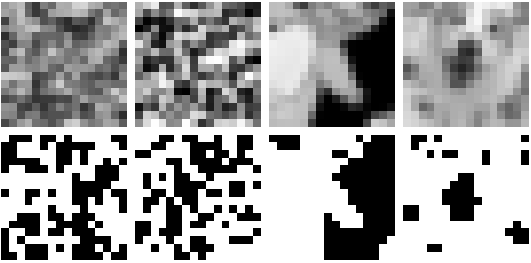}
    \caption{Examples of grayscale data (top) and black and white data (bottom). From the left, Random Noise, Random Dot, Pok\'emon, and ImageNet Dogs. While the randomly generated images are unique, the grayscale and black and white images use the same Pok\'emon and Dog images at similar angles. These images are exemplary of both input and target images in the datasets.}
    \label{fig:datatypes}
\end{figure}

\textbf{Grayscale Data.}
We ran our network on four sets of data, each generated by rotating a base image by the same amount. For both the Pok\'emon and ImageNet Dogs datasets, the first 300 images were used, and random selection was used to select 250 images for the training sets and 50 for evaluation. 

\begin{itemize}
  \item \textit{Random Noise:} randomly generated upscaled 32x32 images where each pixel is a value between 0 and 1
  \item \textit{Random Dot:} randomly generated upscaled 32x32 images where each pixel is 0 or 1
  \item \textit{Pok\'emon:} a collection of 256x256-pixel drawings by Ken Sugimori, selected for their outlines and flat colors \cite{pokemon}
  \item \textit{ImageNet Dogs:} a collection of 300 photographs from the ImageNet Dog synset 
  \cite{imagenet_cvpr09}
\end{itemize}

\begin{figure}[t]
    \centering
    \includegraphics[trim={0 0 0 1.5cm},clip,width=\linewidth]{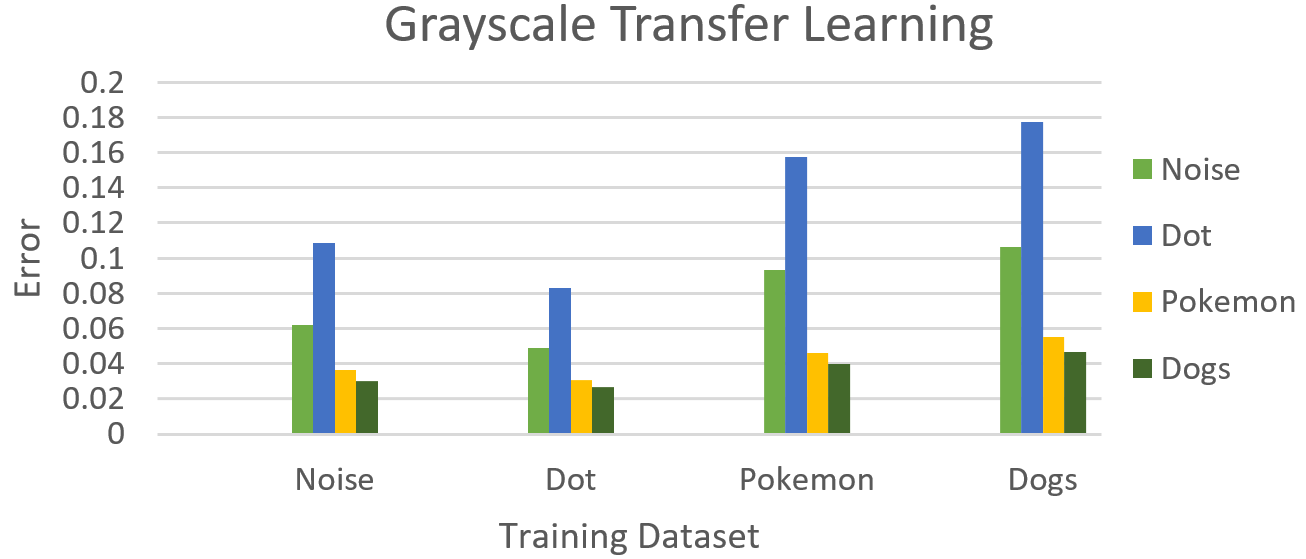}
    \caption{Results from transfer learning using various greyscale datasets.}
    \label{fig:gsTransfer}
\end{figure}

Based on how well each dataset performed when \textit{tested} upon, it is possible to rank them in order of difficulty, with ImageNet Dogs being the easiest, followed by Pokemon, Random Noise, and then Random Dot. The networks \textit{trained} upon these datasets' general performances followed the reverse order. The network trained upon Random Dot, the most difficult dataset, performed the best on all datasets despite reporting poor performance on its training set. 


\textbf{Black and White Data.} 
Given the results of the Grayscale transfer learning, we created a more difficult task. This experiment is identical to the Grayscale Data experiment, but pixels in the 16x16 images have values rounded to 0 or 1.

\begin{figure}[t]
    \centering
    \includegraphics[trim={0 0 0 1.5cm},clip,width=\linewidth]{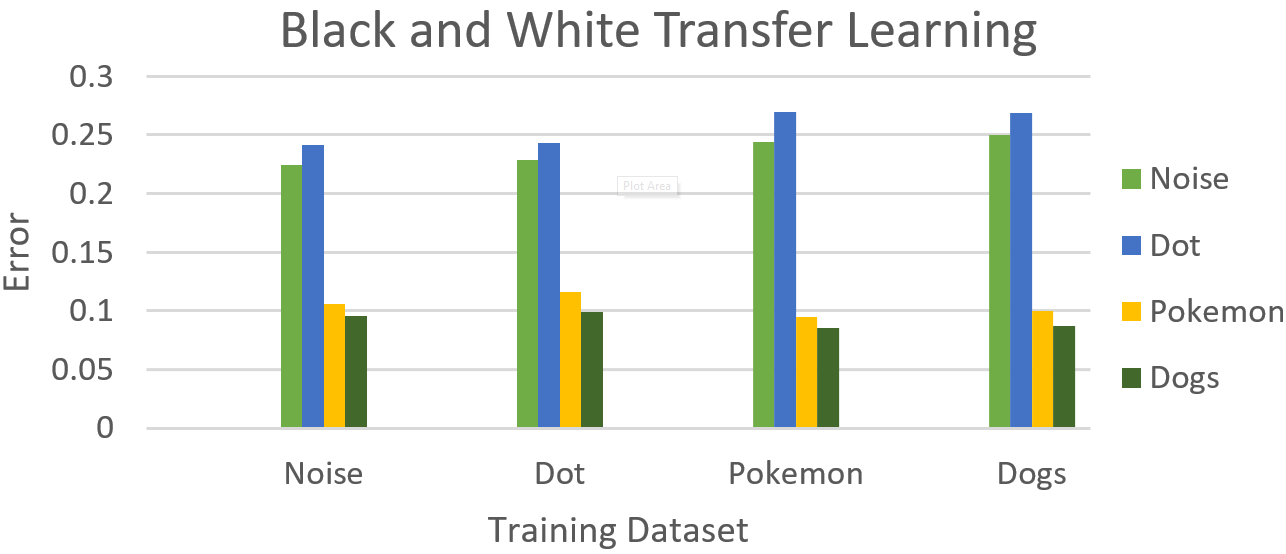}
    \caption{Results from transfer learning using black-and-white datasets.}
    \label{fig:bwTransfer}
\end{figure}

The difficulties of the Black and White datasets follow a noticeably different pattern from the Grayscale difficulties. Random Noise and Random Dot images are both difficult overall, while Pok\'emon and ImageNet Dogs are both easy overall. We see a similar grouping for transfer learning: both of the random datasets perform well on random datasets, and both of the loaded image datasets perform well on loaded images.


\textbf{Transfer between Grayscale and Black and White.}
Finally, we demonstrate the transfer learning between networks trained upon four of the previous datasets: Random Noise (Grayscale), Random Noise (Black and White), Random Dot (Grayscale), and Random Dot (Black and White). 
The network trained on Grayscale Dot images performs the best on all datasets, despite training on the second easiest dataset.

\begin{figure}
    \centering
    \includegraphics[trim={0 0 0 3cm},clip,width=\linewidth]{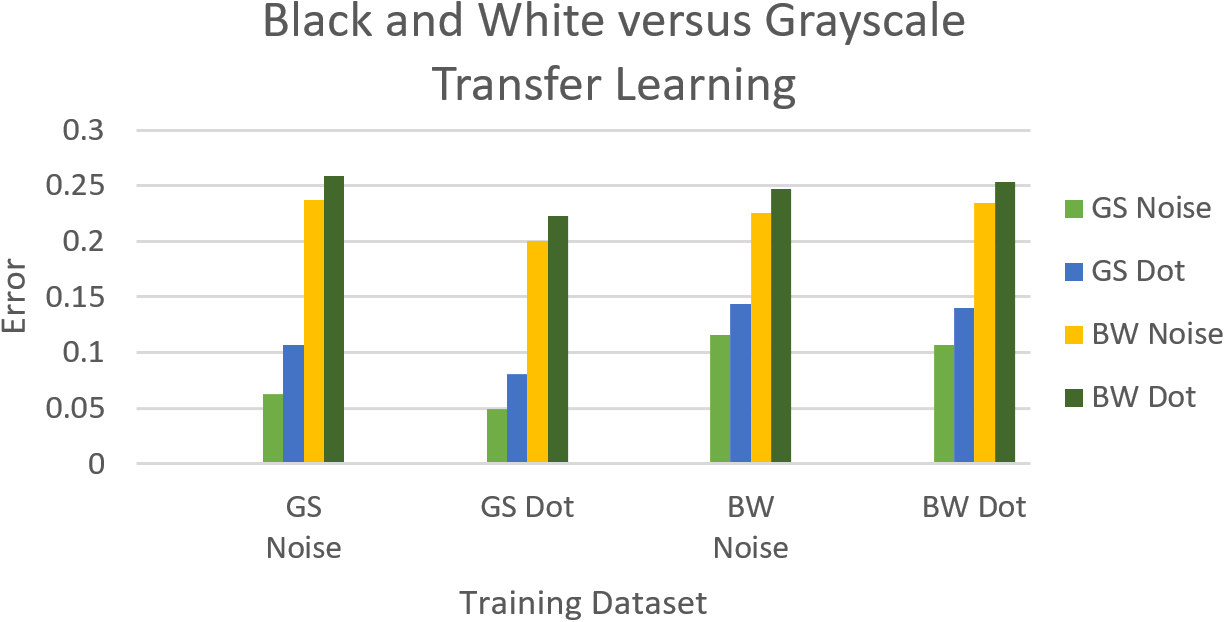}
    \caption{Transfer learning across dataset type and color type.}
    \label{fig:bwgsTransfer}
\end{figure}

\section{Real-World Transformations}

We use videos of handheld objects being rotated and transformed in various ways to demonstrate the network's ability to train on real-world video data. Intra-objecttype and inter-objecttype transfer learning is measured, as well as training on a video of a blank wall (absent).

For each video, 250 frames are used for training, and 50 following frames are used for testing. Specific objects included a ball, a cat, two horses, and a truck. The absent video is only long enough for a training set of 55 images.

\begin{figure}
    \centering
    \includegraphics[width=8.4cm]{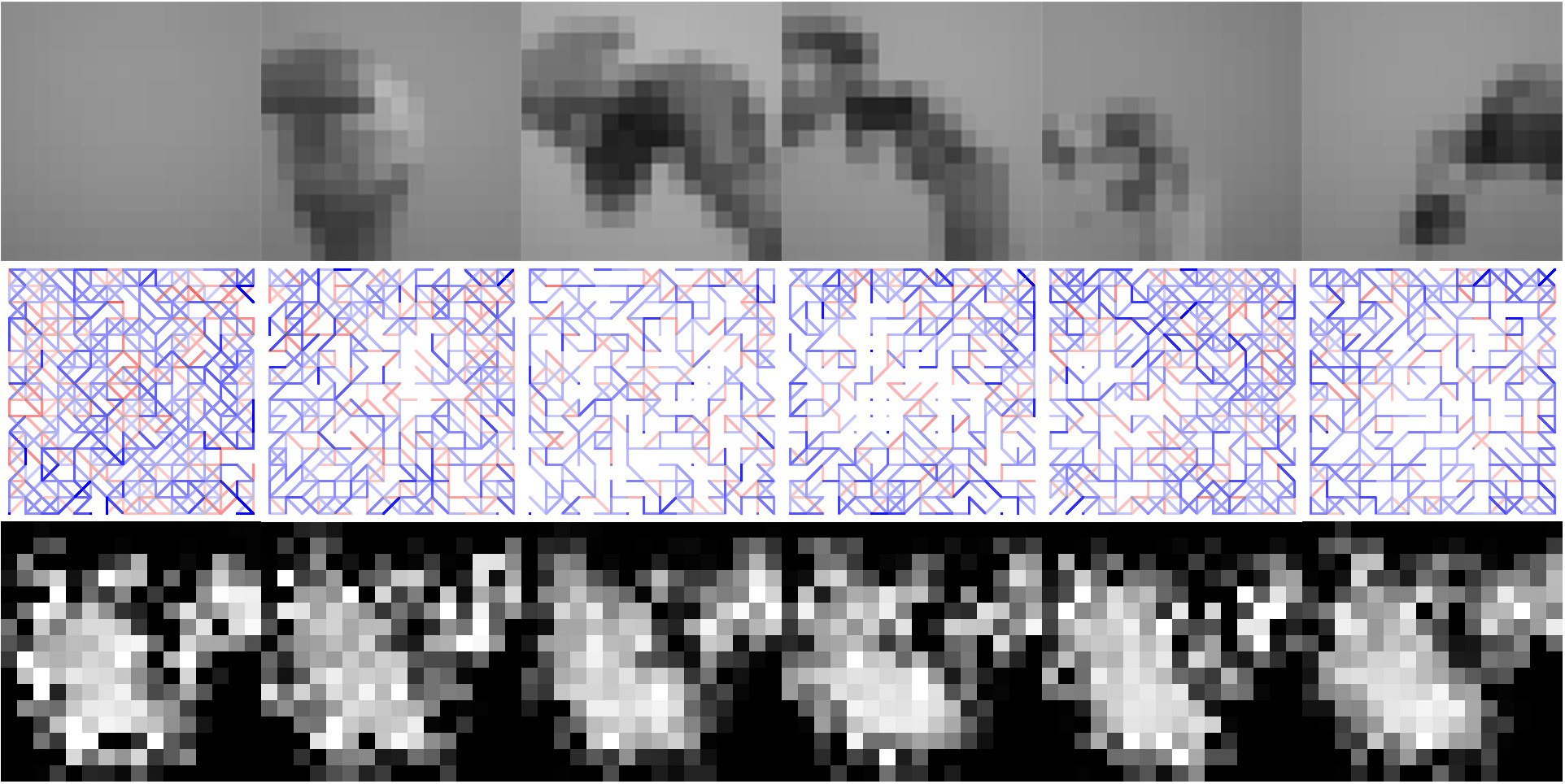}
    \caption{Top: Example images from each training set: absent, ball, cat, horse A, horse B, and truck. Middle: learned structures when training on rotation for the above images. Bottom: 1X-chained output when fed the Pikachu example image. Note that there is no `correct' transformation for these images, though the objects in the videos are rotated counter-clockwise.}
    \label{fig:toybox}
\end{figure}

Because the centers of rotation change, and hands in the video move in varied directions, none of the networks perform well enough to produce recognizable 5X-chained outputs.  Less chaotic regions are observable in the network structures that coincide with object and hand locations in the videos. When objects are large and dark the effect is more pronounced and the example outputs are more clear. Absent forms a surprisingly clear image; all nodes learn to output a color similar to their neighborhoods.

\section{Polar versus Cartesian Networks}

\begin{figure}
    \centering
    \includegraphics[width=8cm]{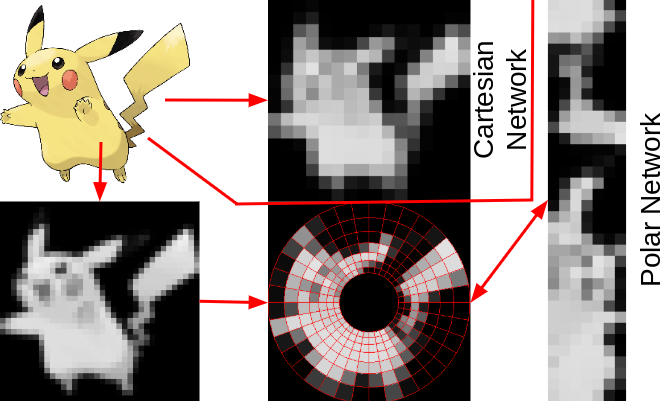}
    \caption{Example of Polar network creation (bottom path) as compared to 16x16 Cartesian network creation (top path). In the former, the original image (top left) is downscaled bilinearly to 36x36. Then it is converted to a 7x36 Polar Picture (far right) that has 7 rings and 36 wedges.}
    \label{fig:polar_demo}
\end{figure}

We now use a network with a polar topology and are especially interested in performance on rotation and scaling, the polar analogues of translation. Our implementation, inspired by early work on circular image representations by Funt \cite{funt_problem-solving_1980}, borrows from an earlier model developed to efficiently rotate images for solving visual reasoning tasks \cite{palmer2018thinking}.
A ``Polar Picture'' consists of square-like regions, called ``sectors''. Each sector is located by a discrete ring (radial coordinate) and wedge (angular coordinate). There is an equal number of sectors in each concentric ring. As radial distance increases, sector size increases and resolution decreases. A central blind spot ensures a finite number of sectors. 
Figure \ref{fig:polar_demo} illustrates polar conversion and compares the two network topologies.

We use a Polar network with 36 wedges to discretize rotation in 10 degree increments. 
We choose 7 rings for the network because 7x36 = 252 total perceptrons, which is close to the Cartesian network's 16x16 = 256. We convert from 36x36 because it yields an innermost sector size of 1.01 pixels, avoiding redundant subpixel regimes.
Prior to downscaling, the Polar pre-processing and transformation pipeline is identical to its Cartesian counterpart. 
We use a dynamic neighborhood radius that varies with the square root of a sector's radial distance to ensure that the larger outermost sectors connect to neighbors.  
Doing so yields 1748 connections, while our Cartesian network has 2116 connections.



We compare translation, rotation, and scaling between Polar and Cartesian networks. The former two use default arguments, (0,1) and 10 respectively. The ratio of sector sizes between neighboring rings is a constant \cite{palmer2018thinking}, in this case 0.839. Our network is thus optimized for a scale factor of 0.839, which we use instead of 0.9.

\begin{figure}
    \centering
    \includegraphics[trim={0 0 0 1.5cm},clip,width=\linewidth]{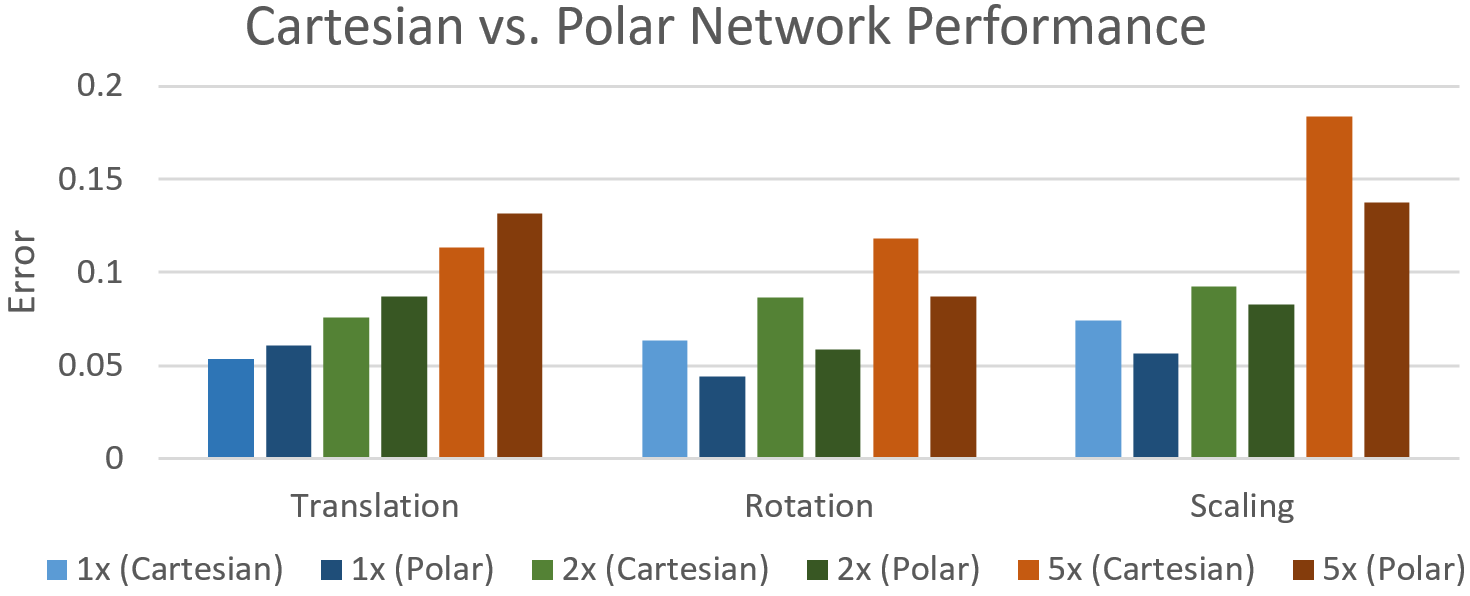}
    \caption{Performance comparison of the two network topologies across the three transformations and chains.}
    \label{fig:cartpolar}
\end{figure}

\begin{figure}
    \centering
    \includegraphics[width=8.4cm]{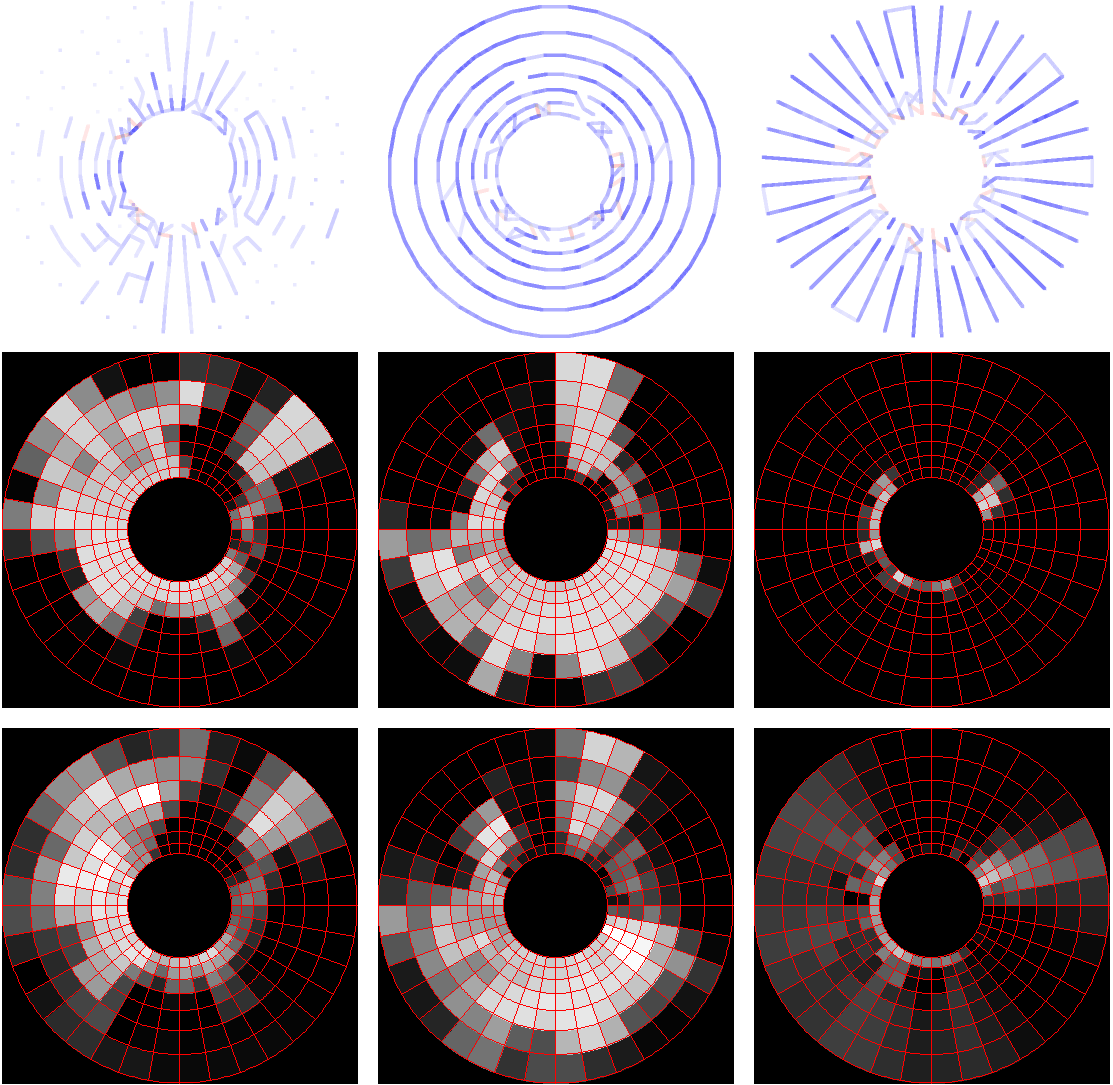}
    \caption{Top: The structure of the networks after training for 5X-chained translation, rotation, and scaling. Middle: The Polar target images for the above tasks. Bottom: The trained networks' outputs for the above tasks. }
    \label{fig:polar}
\end{figure}

As expected, Polar outperforms Cartesian for rotation and scaling in spite of having fewer connections, and Cartesian performs better with translation, seen in Figure \ref{fig:cartpolar}.
Figure \ref{fig:polar} shows how the Polar network 5X chain transforms Pikachus. For comparison, the input Pikachu is bottom middle in Figure \ref{fig:polar_demo}; note the discrete 5X wedge and ring shifts for rotation and scaling. The noise in the scaling network output is similar to that found in Cartesian translation (Figure \ref{fig:baseTranslate}); these outputs are `unlearnable' because no input pixels transform to these areas. 

\section{Conclusion and Future Work}

We presented several experiments to study how image transformations can be represented and learned by planar neural networks---2D spatial layouts of perceptrons.  We showed that basic image transformations like rotation, translation, and scaling, can be learned in a supervised fashion from image sequences, using purely local visual information. Additionally, we characterized the robustness of our learning system along several important dimensions including the parameters of the network and the data it is trained upon.

In future work, we will investigate more complex visual transformations as well as alternate forms of representation, e.g., representing transformations as primitive basis functions instead of discrete entities \cite{goebel1990mathematics}, or developing network structures that more closely emulate the human visual cortex in terms of spatial layout, redundancy, etc. 

Ultimately, we aim to produce a system that learns visuospatial transformations from experience and may flexibly use those transformations to solve new visuospatial reasoning tasks. Although there are architectures that reason using visual transformations \cite{schultheis2011casimir} \cite{wintermute2012imagery}, we are not aware of systems that reason using transformations they have learned from experience.  We continue here in the vein of previous work on learning transformations \cite{mel1986connectionist,memisevic_unsupervised_2007,seepanomwan2013modelling}; 
these kinds of transformation representations could later be integrated into AI architectures for high-level reasoning, and could play an important role in many different tasks ranging from commonsense reasoning to natural language understanding.

\section{Acknowledgments}

This work was funded in part by NSF Award \#1730044.

\bibliographystyle{IEEEtran}
\bibliography{references}

\end{document}